\documentclass[10pt, a4paper]{article}

\usepackage[]{lrec2026} 
\usepackage{amsmath}
\usepackage{comment}
\usepackage{booktabs}
\usepackage{tcolorbox}
\usepackage{caption}
\usepackage{threeparttable}
\usepackage[inline]{enumitem}
\usepackage[table]{xcolor}
\usepackage{amssymb}
\usepackage{bbm}
\usepackage{booktabs}
\usepackage{xspace}

\definecolor{highlight}{RGB}{210, 228, 245}
\newcommand{\evr}{$\text{Ev}^2\text{R}$ }

\title{ClimateCheck 2026: Scientific Fact-Checking and Disinformation Narrative Classification of Climate-related Claims}

\name{Raia Abu Ahmad\textsuperscript{1, 2} \quad Max Upravitelev\textsuperscript{1, 2} \quad Aida Usmanova\textsuperscript{3}\\
  \large \textbf{Veronika Solopova\textsuperscript{1, 2}} \quad \textbf{Georg Rehm\textsuperscript{1, 4}} \\
  }
  
\address{\textsuperscript{1}Deutsches Forschungszentrum für Künstliche Intelligenz GmbH (DFKI), Germany \\
\textsuperscript{2}Technische Universität Berlin, Germany ~~~ \textsuperscript{3}Leuphana Universität Lüneburg, Germany \\
\textsuperscript{4}Humboldt-Universität zu Berlin, Germany \\
\small Corresponding author: \href{mailto:raia.abu_ahmad@dfki.de}{raia.abu\_ahmad@dfki.de}
}

\abstract{
Automatically verifying climate-related claims against scientific literature is a challenging task, complicated by the specialised nature of scholarly evidence and the diversity of rhetorical strategies underlying climate disinformation. ClimateCheck 2026 is the second iteration of a shared task addressing this challenge, expanding on the 2025 edition with tripled training data and a new disinformation narrative classification task. Running from January to February 2026 on the CodaBench platform, the competition attracted 20 registered participants and 8 leaderboard submissions, with systems combining dense retrieval pipelines, cross-encoder ensembles, and large language models with structured hierarchical reasoning. In addition to standard evaluation metrics (Recall@K and Binary Preference), we adapt an automated framework to assess retrieval quality under incomplete annotations, exposing systematic biases in how conventional metrics rank systems. A cross-task analysis further reveals that not all climate disinformation is equally verifiable, potentially implicating how future fact-checking systems should be designed.
 \\ \newline  }

\begin{document}
    
\maketitleabstract

\section{Introduction}

Public understanding of climate change is increasingly shaped by online discourse, the scientific validity of which is often difficult to assess at scale \cite{fownes2018twitter}. While automatic fact-checking (AFC) has made substantial progress in recent years \cite{guo2022survey}, most existing approaches verify claims against general sources such as Wikipedia or web search \cite{thorne-etal-2018-fact,aly-etal-2021-fact,schlichtkrull-etal-2024-automated}. However, recent studies emphasise the importance of using trustworthy evidence sources~\citep{schlichtkrull-etal-2023-intended}, which, for scientific topics, means direct engagement with scholarly literature. 

Although scientific publications constitute one of the most authoritative forms of evidence for climate-related claims, they remain challenging for AFC systems. This stems from several difficulties when dealing with scholarly document processing, such as the use of in-domain terminology, document length, complex reasoning requirements, connection to other documents through citations, and temporal changes of evidence veracity~\cite{vladika-matthes-2023-scientific,deng2025next}.

To tackle this, we introduced the ClimateCheck shared task~\cite{abu-ahmad-etal-2025-climatecheck-shared}, focusing on the verification of climate-related social media claims using scientific abstracts as evidence. The task consists of retrieving relevant abstracts for a given claim, and classifying whether the text \emph{supports}, \emph{refutes}, or does \emph{not have enough information (NEI)} about the claim, treating the problem at the level of individual claim–abstract pairs (CAPs) instead of producing one final verdict. The competition demonstrated the feasibility of grounding AFC in scholarly knowledge, but also exposed several limitations, including restricted training data, evaluation challenges caused by incomplete annotations, and a growing reliance on computationally expensive large language models (LLMs).

This paper presents the 2026 iteration of ClimateCheck, in which we address these challenges and substantially expand the scope of our dataset. We release approx.~three times more training data and introduce a new task, \emph{Disinformation Narrative Classification}, which moves beyond claim-level verification to capture the broader rhetorical structures underlying climate disinformation. 

\begin{figure*}[th!]
    \centering
    \includegraphics[width=0.85\textwidth]{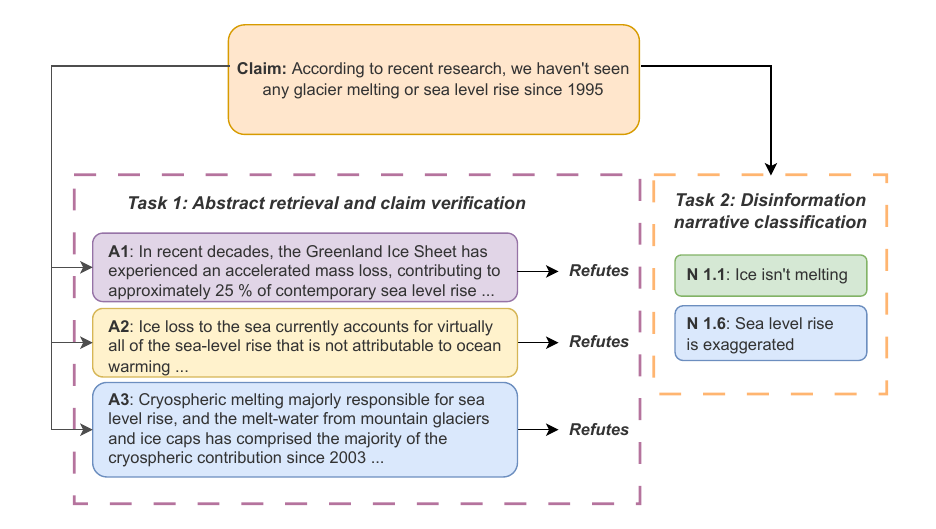}
    \caption{Instance from ClimateCheck 2026. Task 1: Given a claim, systems must retrieve relevant abstracts and use them for verification. Task 2: Given a claim, systems must predict all disinformation narratives associated with it.}
    \label{fig:overview}
\end{figure*}

Crucially, all three tasks: retrieval, verification, and disinformation narrative classification, are annotated over the same set of claims, providing a unified dataset for evidence-grounded narrative classification, as demonstrated in Figure~\ref{fig:overview}.\footnote{Our dataset is publicly available: \url{https://huggingface.co/datasets/rabuahmad/climatecheck}} To the best of our knowledge, this is the first work that enables systematic investigation of how rhetorical narrative structure interacts with evidence retrieval and veracity classification. Our contributions are threefold:

\begin{enumerate}
    \item A unified benchmark connecting scholarly evidence retrieval, claim verification, and narrative classification.
    \item An adapted evaluation framework for incomplete annotations in multi-evidence scientific fact-checking, based on the work proposed by \citet{akhtar2024ev2r}.
    \item An empirical analysis of system behaviour across retrieval, verification, and narrative tasks, highlighting systematic verification biases and persistent challenges in fine-grained narrative classification.
\end{enumerate}

ClimateCheck 2026 ran from January 15 until February 18 on the CodaBench platform~\cite{codabench}, with registration from 20 participants and 8 leaderboard submissions across all tasks. Four teams submitted system descriptions, three of which outperformed our baselines. In this paper, we report our dataset development process (\S\ref{sec:data}), evaluation framework (\S\ref{sec:eval}), baselines design (\S\ref{sec:baselines}), participant systems (\S\ref{sec:results}), and error analyses across tasks (\S\ref{sec:erro}), which we discuss and conclude in \S\ref{sec:discussion} and \S\ref{sec:conclusion}, respectively.

\section{Related Work}
\label{rel-work}

Shared tasks have played a central role in advancing AFC research by establishing common benchmarks and evaluation frameworks. Early efforts such as FEVER~\cite{thorne-etal-2018-fact} and FEVEROUS~\cite{aly-etal-2021-fact} focused on verifying claims against Wikipedia, while subsequent work expanded into more specialised retrieval. SciFact~\cite{wadden-etal-2020-fact,wadden-etal-2022-scifact}, for example, introduced expert-written scientific claims paired with biomedical abstracts, paving the way for the SCIVER shared task~\cite{wadden-lo-2021-overview}, which demonstrated the particular challenges of reasoning over scholarly publications. More recently, AVeriTeC~\cite{schlichtkrull-etal-2024-automated,akhtar-etal-2025-2nd} shifted the focus to real-world claims verified using open-web evidence, highlighting the importance of trustworthy and traceable sources.

As these tasks have grown in complexity, LLM-based systems have come to dominate leaderboards~\cite{akhtar-etal-2025-2nd,schlichtkrull-etal-2024-automated,abu-ahmad-etal-2025-climatecheck-shared}. However, recent work suggests that this trend does not consistently yield superior performance, with smaller or task-specialised models remaining competitive in climate-related settings~\cite{calamai-etal-2025-benchmarking,upravitelev-etal-2025-comparing}.

Claim-level verification, however, addresses only one dimension of the problem. Climate disinformation rarely operates through isolated false claims, as it tends to cluster around recurring narrative patterns that span many individual statements. Several datasets have been developed to capture these patterns, grouping climate denial claims by their underlying narrative messages~\cite{coan2021computer, 2024co, 2025pn}. However, these datasets annotate narrative labels at the claim level without linking them to scientific evidence. To the best of our knowledge, our dataset is the first to connect narrative classification and evidence-based claim verification, enabling the exploration of strategies jointly targeting these tasks.

\section{Task Definitions}
\label{sec:tasks}

The 2026 iteration of ClimateCheck consisted of two tasks: 

\begin{itemize}
    \item \textbf{Task 1: Abstract retrieval and claim verification}: given a social media claim about climate change, retrieve the top 5 most relevant abstracts from the given publications corpus (task~1.1) and classify each claim-abstract pair as \emph{supports}, \emph{refutes}, or \emph{NEI} (task~1.2). 
    
    \item \textbf{Task 2: Disinformation narrative classification}: given a social media claim about climate change, predict which disinformation narrative(s) are present using the schema proposed by \citet{coan2021computer}. 
\end{itemize}

Both tasks were evaluated independently, and participants could choose to submit to either or both. Participants were allowed to use external datasets and apply data augmentation methods.
 
\section{Dataset Development}
\label{sec:data}

Both tasks build on the ClimateCheck dataset introduced in the 2025 iteration~\cite{abu-ahmad-etal-2025-climatecheck}. For task~1, we extend the training split by adding 523 unique English claims, drawn from the original claims pool, and 1897 CAPs, where each claim is linked to 1-5 abstracts. The testing split remains unchanged. The same five annotators who worked on the original dataset also worked on the extension, using the same guidelines described in our previous work. Overall, the dataset for task~1 includes 958 unique claims and 4927 CAPs, each annotated as \emph{supports}, \emph{refutes}, or \emph{NEI}. Table~\ref{tab:label_dist} presents the distribution of labels across the training and test splits. The overall inter-annotator agreement (IAA) of the training split including the extension is Cohen's $\kappa = 0.73$, indicating substantial agreement~\cite{landis1977measurement}.

\begin{table}
\small
\centering
\begin{tabular}{lcc}
\toprule
\textbf{Label} & \textbf{Train} & \textbf{Test}\\
\midrule
Supports & 1399 & 749 \\
Refutes & 451 & 266 \\
Not Enough Information (NEI) & 1173 & 889 \\
\midrule
Total & 3023 & 1904 \\
\bottomrule
\end{tabular}
\caption{Label distribution for Task 1 across training and test splits.}
\label{tab:label_dist}
\end{table}



The same 958 unique claims were annotated for task 2, using disinformation narrative labels from the narrative taxonomy and the code book introduced by~\citet{coan2021computer}. The taxonomy distinguishes 27 climate change denial narrative labels across 5 main narrative groups, plus a no-disinformation label (33 labels in total). Four annotators applied a multi-label scheme to each claim\footnote{Annotation guidelines available at \url{https://github.com/ryabhmd/climatecheck/blob/master/narrative_anno_guide.pdf}}, which we adopted since narratives often overlap, and prior work has shown that, in such cases, single-label annotations can lead to errors and downstream classification failures~\cite{calamai-etal-2025-benchmarking}.
Gold labels were determined by majority vote ($\geq$3/4 agreement), which was achieved for 83.2\% of items (66.6\% unanimous). The remaining 16.8\% were adjudicated by two of the authors (who were not part of the initial annotator pool) through review and discussion.

We report IAA at three levels of granularity, reflecting the hierarchical nature of the annotation scheme. Treating each (item $\times$ label) pair as an independent binary decision yields an overall  Krippendorff's $\alpha = 0.785$. However, this measure 
is dominated by the frequent no-disinformation label (71.5\% prevalence). The prevalence-weighted average of per-label $\alpha$ values, which better reflects agreement on the disinformation narratives themselves, is $\alpha = 0.651$, falling just below the $\alpha \geq 0.667$ threshold for content analysis \citep{krippendorff2018content}. At the level of the five main narrative groups, this  value rises to $\alpha = 0.697$, exceeding the threshold and indicating that annotators identify narratives reliably at the top-level, while finer-grained distinctions prove more challenging. This aligns with studies such as \citet{10706846}, which discuss specific challenges to annotating narrative-related tasks, such as having to account for subjectivity in an interpretive setting. We discuss IAA in further detail and present all labels of the taxonomy in Appendix~\ref{app:iaa}.

\section{Evaluation Process}
\label{sec:eval}

\subsection{Official Ranking Metrics}

\paragraph{Task 1.1: Abstract Retrieval.}
We evaluate retrieval using Recall@K ($K=2,5$) and Binary Preference (Bpref). An abstract is considered evidentiary if it is annotated as \emph{supports} or \emph{refutes} in the gold data.\footnote{More detailed definitions of the metrics are available in Appendix~\ref{app:eval-metrics}.} The final ranking score for task~1.1 is:

\[
\text{Score}_{1.1} =
\frac{1}{2} \left( \text{Recall@5} + \text{Bpref} \right).
\]


\paragraph{Task 1.2: Claim Verification.}
For manually annotated CAPs, we compute weighted precision, recall, and F1. Only annotated pairs are considered; unjudged abstracts are ignored. To ensure that verification performance reflects successful evidence retrieval,  the final ranking score is defined as:

\[
\text{Score}_{1.2} = \text{F1} + \text{Recall@5}.
\]

We add R@5 from task~1.1 to penalise systems that achieve a high verification score on a small number of retrieved pairs.

\paragraph{Task 2: Disinformation Narrative Classification.} Predictions are evaluated against gold claim labels using macro precision, macro recall, and F1 (macro, micro, and weighted). Due to class imbalance, final rankings are determined by macro F1.

\subsection{Evaluation under Incomplete Annotations}

Participant systems may retrieve relevant abstracts that are not annotated in the gold data. It is impossible to annotate every CAP, but we still want to reduce bias towards the retrieval approach we based the gold data on. Thus, to evaluate unannotated CAPs, we adapt the \evr framework \citep{akhtar2024ev2r}, which has proven to have good correlation with human judgments. Due to limited computational resources, \evr scores are computed using the top submission from each team and are reported separately from the official leaderboard.\footnote{We make our implementation of the framework public for future use: \url{https://github.com/ryabhmd/climatecheck/tree/master/automatic_eval}}

The \evr score combines two complementary components: a reference-based component and a proxy-reference component. The first decomposes both retrieved and reference evidence into atomic facts and evaluates their alignment via precision and recall, measuring how accurately and completely the retrieved evidence reflects the reference. The second uses a fine-tuned DeBERTa model~\cite{He21deberta} to predict the veracity label for a given claim–evidence pair, with the model's confidence in the gold label serving as a proxy signal for evidence quality. The final score is a weighted combination of the reference-based F1 score and the proxy-reference confidence score.

\paragraph{Adapted \evr Score.} The original score assumes a single gold reference per claim, which does not hold in our setting since claims may be linked to multiple gold abstracts with evidence-dependent labels. For a claim $c$ with gold abstracts $G_c$ and a retrieved abstract that is unannotated in the gold data $r$, we iterate over $G_c$ and retain the gold abstract with the highest alignment to $r$ using the reference-based component:
\[
S_{\text{ref}}(r) = \max_{g \in G_c} F1(r,g).
\]
Let $g^*$ denote this best-aligned abstract and $y^*$ its label. We then compute the proxy-reference component:

\[
S_{\text{proxy}}(r) = P_\theta(y^* \mid c,r),
\]

measuring how strongly $r$ supports the gold label of the best-aligned abstract. The final adapted \evr score per $r$ is:
\[
\text{\evr}(r) = \frac{1}{2}(S_{\text{ref}}(r) + S_{\text{proxy}}(r)).
\]

For each claim, we take the maximum $\text{\evr}(r)$ across all retrieved abstracts, and the submission-level score is the mean over all claims with at least one non-NEI gold abstract (163 of 172 claims in the test set). Claims linked exclusively to NEI abstracts are excluded from this evaluation as they provide no evidential reference. We discuss our implementation in more depth in Appendix~\ref{app:ev2r}.

\paragraph{Automatic Verification of Unannotated Pairs.}
For retrieved CAPs outside the gold annotations, we additionally evaluate the predicted label $\hat{y}$ based on the proxy component of the \evr framework. We compute a confidence score using the same DeBERTa model:
\[
S_{\text{conf}} = P_\theta(\hat{y} \mid c,r),
\]
and a consistency score against the label of the best-aligned gold abstract:
\[
S_{\text{cons}} = 1[\hat{y} = y^*].
\]
The automatic verification score is then:
\[
S_{\text{ver}} =  \frac{1}{2}(S_{\text{conf}} + S_{\text{cons}})
\]
providing an automatic estimate of label plausibility and consistency for unannotated CAPs, allowing us to assess label quality for retrieved abstracts beyond the annotated gold set.

\section{Baselines}
\label{sec:baselines}

\paragraph{Task 1: Abstract Retrieval and Claim Verification.}
The baseline for task~1 is based on the EFC submission from the 2025 iteration~\cite{upravitelev-etal-2025-comparing}, chosen due to its strong performance relative to its computational requirements. The retrieval component uses a multi-stage pipeline: Abstracts are first retrieved using BM25 \cite{Robertson2009bm25}. The top-$k=1500$ are then re-ranked using semantic similarity computed with a fine-tuned E5 model\footnote{\url{https://huggingface.co/intfloat/e5-large-v2}}, which was trained for 3 epochs on our dataset using contrastive learning. Finally, a MiniLM-based cross-encoder\footnote{\url{https://huggingface.co/cross-encoder/ms-marco-MiniLM-L12-v2}} is applied for re-ranking the top-$k=150$ results, taking the top 5 abstracts per claim as the final predictions. 
For task~1.2, we use the DeBERTa sequence classification model, fine-tuned on six Natural Language Inference (NLI) datasets: MultiNLI~\cite{multinli}, ANLI~\cite{nie-etal-2020-adversarial}, LingNLI~\cite{parrish-etal-2021-putting-linguist}, WANLI~\cite{liu-etal-2022-wanli}, FEVER NLI \cite{nie2019combining}, and the ClimateCheck training split. We optimise for highest minimum accuracy per label to mitigate class imbalance. 

\paragraph{Task 2: Disinformation Narrative Classification.} The baseline employs Qwen3-8B~\cite{qwen3} fine-tuned in a supervised instruction-following setup. Qwen3 is chosen due to good performance on a public leaderboard,\footnote{\url{https://huggingface.co/spaces/k-mktr/gpu-poor-llm-arena}} specifically targeting the evaluation of smaller and more efficient LLMs. The training data is constructed by converting claim–narrative pairs into chat-style instruction–response examples using the CARDS taxonomy (see prompt in Appendix~\ref{app:narr_prompt}). Fine-tuning is performed using parameter-efficient Low-Rank Adaptation \cite[LoRA,][]{hu2022lora} with standard cross-entropy loss on the assistant responses. The model predicts one or more taxonomy codes per claim in a constrained generation format. 

\vspace{4pt}

Our code for implementing baseline models for both tasks is publicly available to enable reproducibility.\footnote{\url{https://github.com/XplaiNLP/ClimateCheck-2026-Baseline/}}

\section{Submitted Systems and Results} 
\label{sec:results}

We received 8 leaderboard submissions across both tasks, the results of which are shown in Table~\ref{tab:results1} for task~1 and Table~\ref{tab:results2} for task~2. We also report scores using the \evr framework for task~1 submissions in Table~\ref{tab:evr}. In what follows, we briefly describe the system descriptions we received from four teams.

\begin{table*}[h]
\centering
\small
\definecolor{highlight}{RGB}{210, 228, 245}
\begin{tabular}{@{}lcccccccc@{}}
\toprule
 & \multicolumn{4}{c}{Task 1.1} & \multicolumn{4}{c}{Task 1.2} \\
\cmidrule(lr){2-5} \cmidrule(lr){6-9}
\textbf{Team} 
& R@2 & R@5 & Bpref & Score$_{1.1}$ 
& P & R & F1 & Score$_{1.2}$ \\
\midrule
Baseline & 0.213 & 0.403 & 0.459 & 0.431 & 0.683 & 0.682 & 0.679 & 1.082 \\
\midrule
Ant Bridge~\cite{wang-etal-2025-winning}$^\dagger$ 
    & \cellcolor{highlight}0.218 
    & \cellcolor{highlight}0.451 
    & \cellcolor{highlight}0.495 
    & \cellcolor{highlight}0.473 
    & \cellcolor{highlight}0.729 
    & \cellcolor{highlight}0.726 
    & \cellcolor{highlight}0.725 
    & \cellcolor{highlight}1.176  \\
\midrule
ClimateSense~\cite{climatesense}
    & \cellcolor{highlight}\textbf{0.221} 
    & \cellcolor{highlight}\textbf{0.443} 
    & \cellcolor{highlight}\textbf{0.489} 
    & \cellcolor{highlight}\textbf{0.466} 
    & \cellcolor{highlight}\textbf{0.742} 
    & \cellcolor{highlight}\textbf{0.744} 
    & \cellcolor{highlight}\textbf{0.740} 
    & \cellcolor{highlight}\textbf{1.183} \\
berkbubus    
    & 0.206 
    & \cellcolor{highlight}0.429 
    & 0.453 
    & \cellcolor{highlight}0.441 
    & 0.391 
    & 0.414 
    & 0.328 
    & 0.757 \\
DFKI-IML~\cite{dfki-iml}     
    & 0.193 & 0.352 & 0.401 & 0.377 
    & 0.670 & 0.641 & 0.620 & 0.972 \\
gardlz       
    & 0.193 & 0.364 & 0.314 & 0.340 
    & 0.581 & 0.593 & 0.579 & 0.943 \\
ytsoneva     
    & 0.060 & 0.094 & 0.116 & 0.105 
    & 0.269 & 0.176 & 0.148 & 0.242 \\
\bottomrule
\end{tabular}
\begin{tablenotes}
\small
\item $^\dagger$ Winning team of the 2025 iteration on the same test set; included for reference only.
\end{tablenotes}
\caption{Results for Task 1. Task 1.1 is evaluated using Recall@2 (R@2), Recall@5 (R@5), Binary Preference (Bpref), and Score$_{1.1}$. Task 1.2 is evaluated using weighted Precision (P), Recall (R), F1, and Score$_{1.2}$. Best results among 2026 submissions are shown in \textbf{bold}. Shaded cells indicate scores that outperform the baseline.}
\label{tab:results1}
\end{table*}

\begin{table*}[h]
\centering
\small
\begin{tabular}{@{}lccccc@{}}
\toprule
\textbf{Team} 
& P (macro) 
& R (macro)
& F1 (macro)
& F1 (micro)
& F1 (weighted) \\
\midrule 
Baseline & 0.530 & 0.574 & 0.514 & 0.798 & 0.784 \\
\midrule
ahilbert~\cite{ahilbert} 
    & \cellcolor{highlight}\textbf{0.707} 
    & \cellcolor{highlight}0.631
    & \cellcolor{highlight}\textbf{0.625} 
    & \cellcolor{highlight}0.844 
    & \cellcolor{highlight}0.821 \\
XplaiNLP~\cite{xplainlp} 
    & \cellcolor{highlight}0.625 
    & \cellcolor{highlight}\textbf{0.640} 
    & \cellcolor{highlight}0.597 
    & \cellcolor{highlight}0.801 
    & \cellcolor{highlight}0.806 \\
ClimateSense~\cite{climatesense} 
    & \cellcolor{highlight}0.670 
    & 0.568 
    & \cellcolor{highlight}0.583 
    & \cellcolor{highlight}\textbf{0.876} 
    & \cellcolor{highlight}\textbf{0.844} \\
alextsiakalou 
    & \cellcolor{highlight}0.574 
    & \cellcolor{highlight}0.586 
    & \cellcolor{highlight}0.548 
    & \cellcolor{highlight}0.844 
    & \cellcolor{highlight}0.837 \\
\bottomrule
\end{tabular}
\caption{Results for Task 2. Systems are evaluated using macro-averaged Precision (P), Recall (R), and F1, as well as micro- and weighted F1. Macro-F1 is the official ranking metric, reflecting performance across all narrative categories regardless of class frequency. Best results are shown in \textbf{bold}. Shaded cells indicate scores that outperform the baseline.}
\label{tab:results2}
\end{table*}

\paragraph{ClimateSense~\cite{climatesense}.} For task~1, the team employed a three-stage pipeline, following the winning team of the 2025 iteration~\cite{wang-etal-2025-winning}. Initial candidate retrieval used BM25, followed by an ensemble of five fine-tuned BGE re-rankers~\cite{chen-etal-2024-m3} trained with hard negative mining, and aggregated via Reciprocal Rank Fusion. For claim verification, the team opted for zero-shot (ZS) classification using gpt-oss-120b\footnote{\url{https://huggingface.co/openai/gpt-oss-120b}}. A fine-tuned DeBERTa-based cross-encoder was also explored but underperformed compared to the ZS LLM approach. For task~2, they adopted a ZS approach using GPT 5.1\footnote{\url{https://openai.com/index/gpt-5-1/}}. The prompt incorporated the full CARDS taxonomy definitions, negative criteria for each category, specificity guidelines, and five-step chain-of-thought (CoT) reasoning. 

\paragraph{DFKI-IML~\cite{dfki-iml}.} The team focused mainly on task~1.2, adopting a similar retrieval pipeline as the baseline and directing their attention to claim verification. They compared two self-explainable inference paradigms: Intermediate Reasoning (IR), where the model generates an explanation before predicting the entailment label, and Post-hoc Rationalization (PR), where the label is predicted before the explanation. Three models were evaluated under both paradigms in a ZS setting: GPT-4o-mini~\cite{achiam2023gpt}, Phi-3-Medium~\cite{abdin2024phi3techreport}, and Mistral-Nemo\footnote{\url{https://huggingface.co/mistralai/Mistral-Nemo-Instruct-2407}}. IR yielded more stable and consistent results across all three models, with GPT-4o-mini achieving the highest scores. 

\paragraph{ahilbert~\cite{ahilbert}.} The team focused exclusively on task~2, using Qwen3-8B as a fixed backbone to compare approaches. They investigated three directions: data augmentation, prompt engineering, and reinforcement learning via Group-Relative Policy Optimization~\cite[GRPO,][]{shao2024deepseekmath}. They augmented the data with synthetic claims, applying skewed inverse-frequency reweighting to oversample minority narratives. For prompt engineering, they compared three strategies: a simple direct instruction prompt, a hierarchical two-turn approach, and a CoT prompt that encourages the same hierarchical reasoning within a single forward pass through claim decomposition, high-level group selection, and sub-label assignment. The best-performing setup was ZS CoT prompting combined with Qwen3's native reasoning mode. The team also reported inference-time emissions, showing a clear trade-off between reasoning-based performance and computational cost, with reasoning-heavy approaches consuming roughly 25 times more energy than ZS inference.

\paragraph{XplaiNLP~\cite{xplainlp}.} The team worked on task~2, investigating both encoder-based and decoder-only approaches. For the former, they fine-tuned ModernBERT-large~\cite{Warner25ModernBERT}, RoBERTa~\cite{liu2019roberta}, and DistilBERT~\cite{sanh2019distilbert} on the the training data, additionally using the Augmented CARDS dataset~\cite{rojas24cards} to mitigate class imbalance. ModernBERT with the augmented dataset achieved the strongest results, surpassing our baseline at a fraction of the computational cost. For decoder-only models, the team fine-tuned Qwen3-8B under three instruction tuning strategies: prompt enhancement with in-context examples, hierarchical instruction tuning where top-level and fine-grained categories were predicted in two sequential stages, and retrieval-augmented instruction tuning where a cross-encoder first retrieved the ten most semantically similar narrative descriptions from the CARDS taxonomy. The retrieval-augmented approach consistently outperformed the other strategies. 

\begin{table*}[h]
\centering
\small

\begin{tabular}{@{}lcccccc@{}}
\toprule
\textbf{Team} 
& $\text{Ev}^2\text{R}$(r) 
& $S_{\text{ver}}$ 
& \# Evaluated 
& Avg.\ Unannotated \\
& & & Claims & Abstracts per Claim \\
\midrule
Baseline      & 0.606 & 0.600 & 112 & 1.92 \\
\midrule
ClimateSense~\cite{climatesense}  & 0.559 & 0.559 & 111 & 1.77 \\
berkbubus     & 0.554 & 0.468 & 106 & 1.67 \\
DFKI-IML~\cite{dfki-iml}      & 0.568 & 0.527 & 135 & 1.97 \\
gardlz        & 0.585 & 0.528 & 134 & 2.16 \\
ytsoneva      & 0.672 & 0.143 & 163 & 4.06 \\
\bottomrule
\end{tabular}
\caption{\evr evaluation results for unannotated abstracts retrieved in Task 1. $\text{Ev}^2\text{R}$(r) is the mean of the reference-based and proxy-reference components evaluating task~1.1. $S_{\text{ver}}$ is the added verification score evaluating task~1.2. Only the final leaderboard submission per team is evaluated.}
\label{tab:evr}
\end{table*}

\section{Error Analysis}
\label{sec:erro}

\begin{figure}[th!]
\begin{center}
\includegraphics[width=0.95\columnwidth]{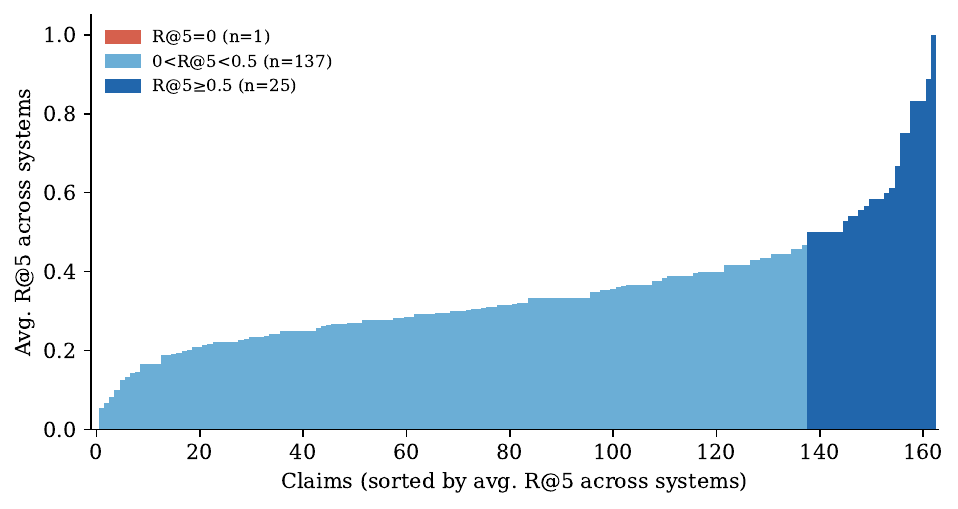}
\captionof{figure}{Per-claim average Recall@5 across all submitted systems, sorted in ascending order. Colours indicate retrieval difficulty.}
\label{fig:retrieval-diff}
\end{center}
\end{figure}

\paragraph{Retrieval Difficulty. } Figure~\ref{fig:retrieval-diff} shows the distribution of per-claim average Recall@5 across all submitted systems, sorted in ascending order. The distribution is notably gradual, suggesting that retrieval difficulty is not a binary easy/hard problem. Only one claim achieves Recall@5=0 across all systems,\footnote{ ``NOAA has adjusted past temperatures to look colder and recent ones warmer''} despite being linked to evidentiary abstract in the gold data. A possible reason could be a lexical mismatch between the claim and the available abstracts, preventing sparse retrieval systems (i.e. BM25, used by all teams) from surfacing relevant evidence. The vast majority of claims (n=137) fall in the mid-range between 0 and 0.5, and the easiest claims (n=25, R@5$\geq$0.5) are rarely solved perfectly by all systems, indicating that no system consistently achieves high recall, even on claims where retrieval is easiest. 

\begin{figure*}[th!]
    \centering
    \includegraphics[width=0.8\textwidth]{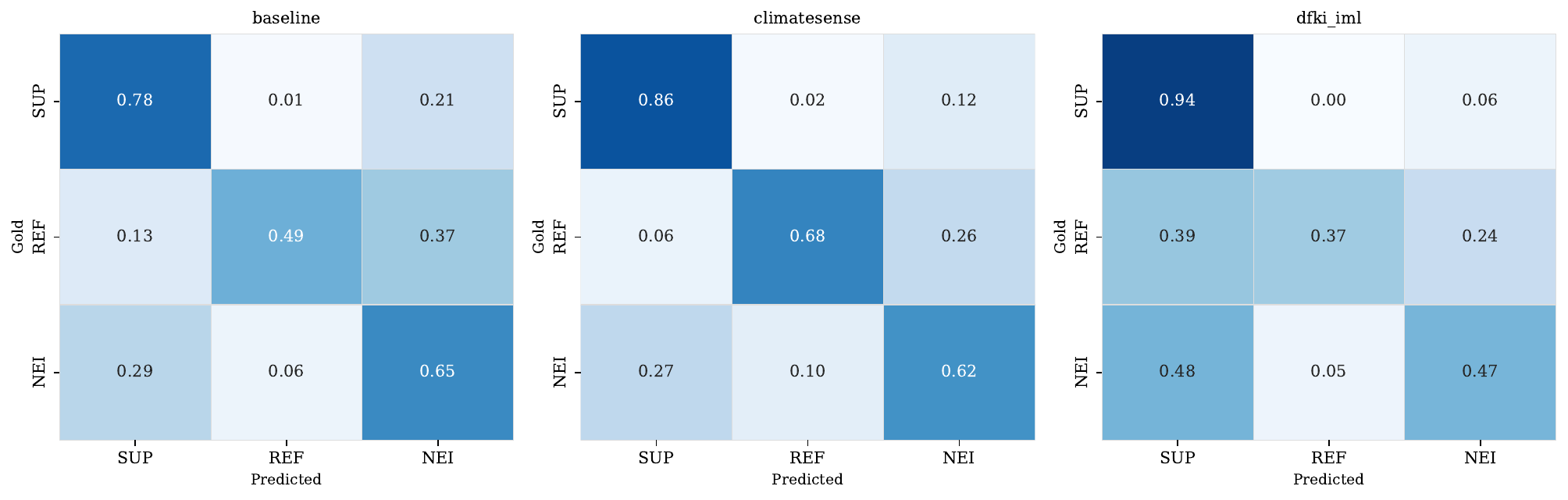}
    \caption{Confusion matrices for the baseline, ClimateSense, and DFKI-IML predictions on task~1.2, normalised by claim. SUP = Supports, REF = Refutes, NEI = Not Enough Information.}
    \label{fig:confusion_matrices}
\end{figure*}

\paragraph{Verification Label Confusion. } Figure~\ref{fig:confusion_matrices} shows normalised confusion matrices for submitted systems from our baseline, ClimateSense, and DFKI-IML on task~1.2.\footnote{We show further systems, whose teams did not submit reports, in Appendix~\ref{app:error-analysis}.} A clear pattern emerges across systems: \emph{Refutes} is the hardest label to predict correctly, with diagonal values ranging from 0.05 to 0.68. This is consistent with the broader NLI literature, showing that refutation requires stronger evidence-grounded reasoning than support or abstention~\cite{Atanasova22, Thorne18Fever}. The most common error is the confusion of \emph{NEI} with \emph{Supports}: the baseline misclassifies 29\% of such instances, ClimateSense 27\%, and DFKI-IML 48\%. This suggests that systems tend to interpret partial or weakly relevant evidence as supporting rather than insufficient. 
We also note a particular error pattern in DFKI-IML, where 39\% of gold \emph{Refutes} instances are predicted as \emph{Supports}, meaning that the system flips its interpretation of refuting evidence, potentially leading to false affirmation of misinformation.

\begin{figure}[h]
  \centering
  \includegraphics[width=0.95\columnwidth]{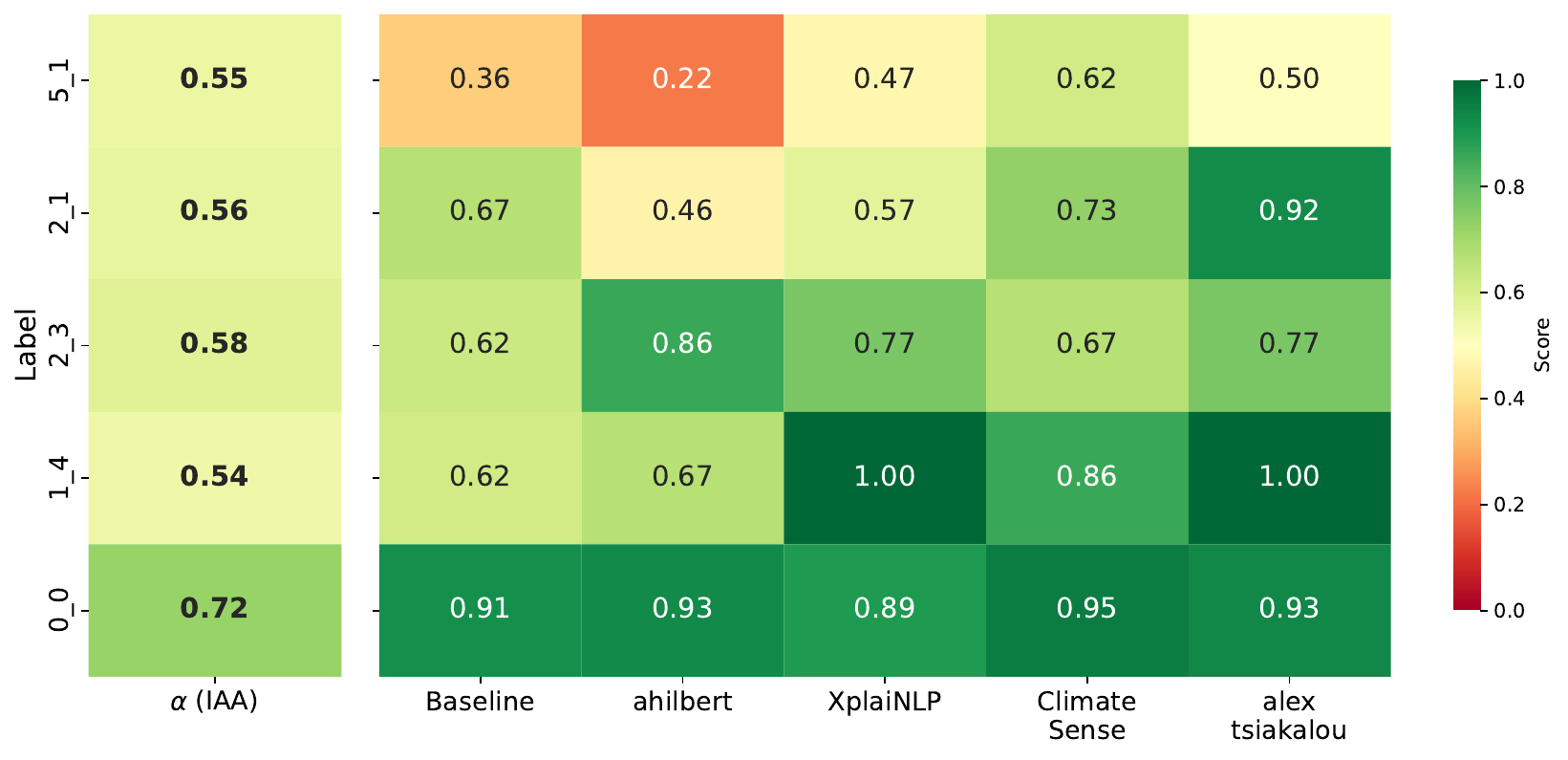}
  \caption{Per-label F1 scores across participating systems and IAA (Krippendorff's $\alpha$, leftmost column). Labels sorted by mean system F1 (ascending); only labels with more than 3 instances in the test set are shown.}
  \label{fig:heatmap_iaa}
\end{figure}

\paragraph{Narrative Classification.} We analyse error patterns of the submitted systems on Task 2. Figure~\ref{fig:heatmap_iaa} shows per-label F1 scores for each system alongside IAA (Krippendorff's $\alpha$). On average, 80.9\% of test claims received an exact-match prediction, 3.5\% were \emph{partially correct} (correct top-level category, wrong sub-narrative), and 15.6\% were assigned a completely wrong top-level category. Of wrong predictions, 81\% crossed top-level category boundaries while only 19\% stayed within the correct category, indicating that identifying the broad narrative type is the primary challenge rather than distinguishing sub-narratives. Systems also showed a consistent under-prediction tendency: 4.8\% of predictions contained too few labels versus only 0.6\% with too many, with 0\_0 (\emph{no disinformation}) being the most common false prediction. Labels that were difficult for annotators were also difficult for systems. Among labels with sufficient test support, Krippendorff's $\alpha$ and mean system F1 show a perfect rank correlation (Spearman $\rho{=}1.0$, $n{=}4$), though the small number of qualifying labels limits statistical power. This pattern is visible in Figure~\ref{fig:heatmap_iaa}: label 5\_1 ($\alpha{=}0.55$) has both the lowest and most variable system F1 (0.22-0.62), while 0\_0 ($\alpha{=}0.72$) is consistently well-handled (F1~${\geq}\,0.89$). Label 2\_1 (\emph{natural cycles}; $\alpha{=}0.56$) shows the widest inter-system spread (0.46--0.92), suggesting that system architecture matters most for labels with moderate annotator agreement. Seven claims (4.1\%) were misclassified by all five systems (F$_1{=}0.0$), and a majority-vote ensemble scored below the best individual system (84.3\% vs.\ 85.5\% exact match), indicating correlated errors.

\begin{figure}[th!]
\begin{center}
\includegraphics[width=0.95\columnwidth]{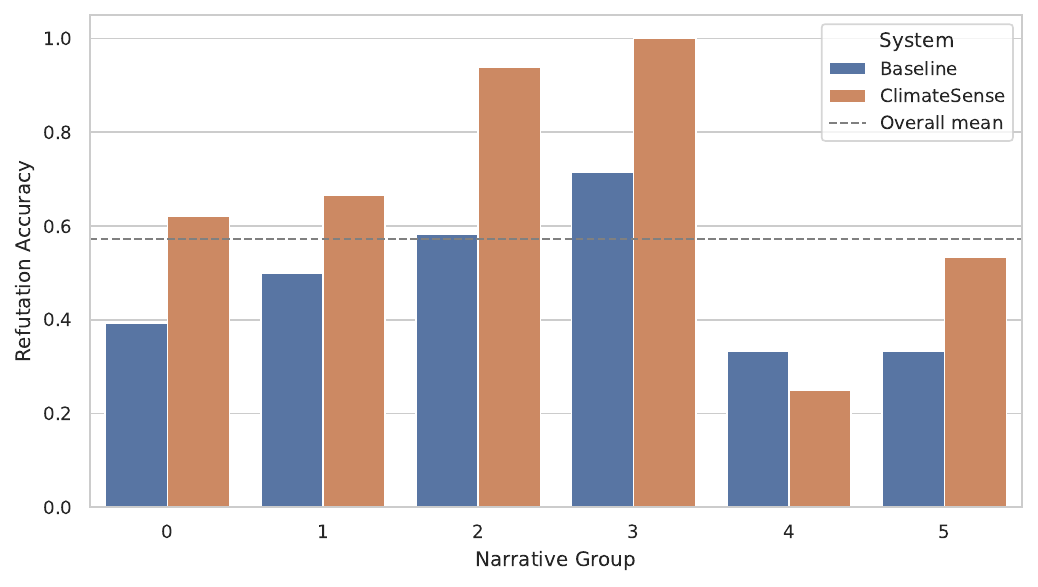}
\captionof{figure}{Refutation accuracy by narrative group based on the CARDS taxonomy for the baseline and ClimateSense systems.}
\label{fig:cross-task}
\end{center}
\end{figure}

\paragraph{Cross-task Analysis.} To investigate whether disinformation narratives affects retrieval and verification difficulty, we analyse system performance across the five main narrative groups using the two systems that submitted predictions for both tasks: our baseline and ClimateSense. We first note that retrieval is narrative-agnostic, with Recall@5 results being consistently high across all groups.\footnote{See Appendix~\ref{app:error-analysis} for full results.} This indicates that relevant scientific evidence is retrievable regardless of the type of disinformation narrative a claim embodies. However, we see a different pattern when looking at verification difficulty, signaling that it is strongly narrative-dependent. Looking at the most difficult label to verify, \emph{Refutes}, we note a clear gradient where claims belonging to Group 3 (\emph{climate impacts are not bad}) are the easiest to refute, while Groups 4 (\emph{climate solutions won't work}) and 5 (\emph{climate movement/science is unreliable}) are systematically the hardest (See Figure~\ref{fig:cross-task}). This result is structurally motivated, since Group 4 claims are primarily normative or economic in nature, and Group 5 claims are epistemological, attacking the credibility of science itself rather than making falsifiable empirical assertions. Thus, using a scientific abstract to refute a claim that science is unreliable is self-referentially problematic: the evidence source is precisely what the claim calls into question. This represents a structural limitation of evidence-grounded AFC that is difficult to resolve through improved retrieval or stronger verification models alone.

\section{Discussion}
\label{sec:discussion}

\paragraph{System Trends.} Results across both tasks reveal a consistent pattern: structured reasoning and task-specific architectural choices matter more than model size or data volume alone. For task~1, the strongest retrieval systems rely on multi-stage dense re-ranking pipelines with cross-encoder ensembles, echoing findings from the 2025 iteration. Notably, although ClimateSense adopts a retrieval approach similar to the 2025 winning team, their retrieval performance remains lower despite the availability of more training data in this iteration, suggesting that implementation details and training methodology are more important than data volume at this scale. For task~2, the most effective approaches share a common thread: hierarchical or structured reasoning that identifies coarse-grained narrative groups before committing to fine-grained sub-labels. Team ahilbert's results further highlight an important practical consideration, raising questions about the deployability of LLM-based verification at scale in real-world AFC pipelines.

\paragraph{Evaluation Beyond Gold Annotations.} The \evr results reveal trends that the official leaderboard alone does not capture. Most notably, team \emph{ytsoneva}, which ranks last on the official retrieval metric, achieves the highest $\text{Ev}^2\text{R}$(r) score. This is largely explained by its retrieval behaviour: it is the only system that retrieves at least one unannotated abstract for all 163 evaluated claims, with an average of more than 4 out of 5 retrieved abstracts per claim falling outside the gold annotations. This means the official metrics penalise this system almost entirely for retrieving evidence that was never judged, rather than for retrieving poor evidence. The \evr score suggests that this evidence is in fact of reasonable quality, a signal that wouldn't be visible under standard evaluation alone. Beyond this case, the baseline achieves the highest \evr score among the remaining systems despite not ranking first on the official leaderboard, further suggesting that systems optimising directly for gold annotations may be learning retrieval biases introduced by the annotation process. Taken together, these results reinforce the importance of complementing recall-based metrics with automatic evaluation frameworks whenever annotation completeness cannot be guaranteed, which is a particularly pressing concern in AFC, where exhaustive annotation is rarely feasible.

\paragraph{The Cross-Task Connection.} Our cross-task analysis reveals that verification difficulty is strongly narrative-dependent, with narrative groups 4 and 5 proving systematically resistant to evidence-based refutation. This finding has a direct implication for system design: a pipeline that first classifies the disinformation narrative of a claim and then conditions its verification strategy on that classification is a natural and promising direction. For claims that make concrete empirical assertions, standard evidence retrieval and NLI-based verification may be sufficient. For other claims, however, the structural mismatch between the claim type and the evidence source suggests that different strategies might be needed, such as specialised evidence types beyond scholarly abstracts, or flagging for human review rather than automated verdict. Crucially, none of the submitted systems attempted to exploit the connection between the two tasks, leaving narrative-aware verification an open and promising direction that can be explored further.

\section{Conclusion}
\label{sec:conclusion}

We presented ClimateCheck 2026, expanding the shared task with tripled training data, a new disinformation narrative classification task, and a unified benchmark connecting evidence retrieval, claim verification, and narrative classification over a common set of claims. Three of four participating teams outperformed our baselines, with multi-stage dense retrieval pipelines and structured hierarchical reasoning proving the most effective strategies for tasks 1 and 2 respectively. Beyond system-level results, our cross-task analysis reveals a structural limitation of evidence-grounded AFC: while retrieval difficulty is narrative-agnostic, verification difficulty is strongly narrative-dependent, with epistemological and normative claims proving more difficult to scholarly refutation. This motivates narrative-aware verification as a promising direction for future work, which our unified benchmark is uniquely positioned to support.

\section{Limitations}

The ClimateCheck dataset is restricted to English-language claims only, limiting the applicability of trained systems to the broader multilingual landscape of online climate discourse. 
In addition, the claim verification task, as formulated here, treats verification as a pairwise problem between a single claim and a single abstract. This is a simplification that enables scalable annotation, but does not reflect the full complexity of scientific AFC, where a verdict may depend on evidence across multiple documents, conflicting findings, and following chains of citations. Future iterations could explore multi-hop verification settings, where evidence from several abstracts must be jointly considered to reach a final verdict.

In terms of participating systems, we note that, for task~1, teams largely replicated or incrementally extended the methods that proved successful in the 2025 iteration of the shared task, with little methodological novelty in the retrieval stage. Additionally, despite all claims being annotated for both verification and disinformation narrative labels, no team explored whether narrative predictions could inform veracity classification or vice versa. Given that disinformation narratives carry implicit expectations about the direction of evidence, narrative-aware verification represents a promising direction.

Finally, the annotation of fine-grained disinformation narratives remains challenging even for human annotators, as reflected in the below-threshold IAA for several categories. This inherent subjectivity sets an upper bound on what automated systems can be expected to achieve, and suggests that future work may benefit from the introduction of hierarchical evaluation metrics that reward partial credit for correct top-level classification.

\section{Acknowledgments}
This work was supported by the consortium NFDI for Data Science and Artificial Intelligence (NFDI4DS)\footnote{\url{https://www.nfdi4datascience.de}} as part of the non-profit association National Research Data Infrastructure (NFDI e.\,V.). The consortium is funded by the Federal Republic of Germany and its states through the German Research Foundation (DFG) project NFDI4DS (no.~460234259). Furthermore, the work was partly performed in the scope of the projects ``VeraXtract'' (reference: 16IS24066) and ``news-polygraph'' (reference: 03RU2U151C) funded by the German Federal Ministry for Research, Technology and Aeronautics (BMFTR). We  thank the annotators: Emmanuella Asante, Farzaneh Hafezi, Senuri Jayawardena, and Shuyue Qu for working on the extension of the task~1 data, and Berk Bubus, Paul-Conrad Feig, Neda Foroutan, Alexandra Tsiakalou for annotating the task~2 data. We also thank Nikolas Rauscher for helping with training the DeBERTa model used for computing the \evr scores.

\clearpage
\section{Bibliographical References}\label{sec:reference}

\bibliographystyle{lrec2026-natbib}
\bibliography{lrec2026-example}


\clearpage

\appendix

\section{Inter-Annotator Agreement And Annotation Process of Task~2}
\label{app:iaa}
\begin{table*}[h]
\centering
\small
\begin{tabular}{@{}lp{7.8cm}rrrrr@{}}
\toprule
\textbf{Label} & \textbf{Narrative} & \textbf{$N$} & \textbf{Prev.} & \textbf{$\alpha$} & \textbf{$\kappa$} & \textbf{PA} \\
\midrule
\multicolumn{7}{@{}l}{\textit{0 --- No disinformation narrative}} \\
0\_0 & No disinformation narrative detected & 776 & .715 & .722 & .722 & .921 \\
\midrule
\multicolumn{7}{@{}l}{\textit{1 --- Global warming is not happening}} \\
1\_0 & Global warming is not happening (general) & 12 & .004 & .175 & .132 & .133 \\
1\_1 & Ice/permafrost/snow cover isn't melting & 28 & .017 & .672 & .676 & .681 \\
1\_2 & We're heading into an ice age/global cooling & 23 & .011 & .556 & .570 & .575 \\
1\_3 & Weather is cold/snowing & 10 & .004 & .350 & .313 & .315 \\
1\_4 & Climate hasn't warmed over the last decade(s) & 36 & .018 & .544 & .536 & .543 \\
1\_5 & Oceans are cooling/not warming & 4 & .002 & .583 & .610 & .611 \\
1\_6 & Sea level rise is exaggerated/not accelerating & 22 & .016 & .795 & .793 & .797 \\
1\_7 & Extreme weather isn't increasing/not linked to CC & 22 & .010 & .427 & .365 & .370 \\
1\_8 & Changed name from `global warming' to `climate change' & 3 & .002 & .750 & .738 & .739 \\
\midrule
\multicolumn{7}{@{}l}{\textit{2 --- Human greenhouse gases are not causing climate change}} \\
2\_0 & Human GHGs are not causing CC (general) & 14 & .004 & .080 & .043 & .044 \\
2\_1 & It's natural cycles/variation & 94 & .051 & .562 & .561 & .583 \\
2\_2 & Non-GHG human forcings (aerosols, land use) & 12 & .003 & $-$.003 & $-$.003 & .000 \\
2\_3 & No evidence for GHG effect driving climate change & 48 & .028 & .581 & .578 & .590 \\
2\_4 & CO$_2$ is not rising/ocean pH is not falling & 3 & .003 & .800 & .750 & .750 \\
2\_5 & Human CO$_2$ emissions are miniscule & 16 & .006 & .273 & .277 & .281 \\
\midrule
\multicolumn{7}{@{}l}{\textit{3 --- Climate impacts/global warming is beneficial/not bad}} \\
3\_0 & Climate impacts are beneficial/not bad (general) & 14 & .005 & .230 & .187 & .189 \\
3\_1 & Climate sensitivity is low/negative feedbacks & 7 & .003 & .265 & .281 & .283 \\
3\_2 & Species/ecosystems aren't impacted/are benefiting & 22 & .014 & .688 & .686 & .690 \\
3\_3 & CO$_2$ is beneficial/plant food & 30 & .015 & .568 & .585 & .591 \\
3\_4 & It's only a few degrees (or less) & 19 & .006 & .190 & .175 & .179 \\
3\_5 & CC doesn't contribute to conflict/threaten security & 8 & .003 & .131 & .073 & .074 \\
3\_6 & CC doesn't negatively impact health & 3 & .002 & .666 & .694 & .694 \\
\midrule
\multicolumn{7}{@{}l}{\textit{4 --- Climate solutions won't work}} \\
4\_0 & Climate solutions won't work (general) & 2 & .001 & $-$.000 & --- & .000 \\
4\_1 & Climate policies are harmful & 13 & .004 & .122 & .100 & .103 \\
4\_2 & Climate policies are ineffective/flawed & 32 & .013 & .255 & .254 & .264 \\
4\_3 & Too hard to solve (politically/economically/technically) & 16 & .006 & .301 & .297 & .301 \\
4\_4 & Clean energy/biofuels won't work & 26 & .010 & .293 & .260 & .266 \\
4\_5 & People need energy from fossil fuels/nuclear & 9 & .004 & .398 & .363 & .365 \\
\midrule
\multicolumn{7}{@{}l}{\textit{5 --- Climate movement/science is unreliable}} \\
5\_0 & Climate movement/science is unreliable (general) & 1 & .000 & .000 & --- & .000 \\
5\_1 & Science is uncertain/unsound (data, methods, models) & 70 & .041 & .551 & .551 & .570 \\
5\_2 & Climate movement is alarmist/political/biased & 26 & .010 & .238 & .214 & .221 \\
5\_3 & Climate change is a conspiracy & 6 & .002 & .249 & .261 & .262 \\
\bottomrule
\end{tabular}
\caption{Per-label IAA sorted by top-level category. $N$ = items where $\geq$1 annotator assigned the label; Prev.\ = prevalence; $\alpha$ = Krippendorff's alpha; $\kappa$ = avg.\ pairwise Cohen's kappa; PA = positive agreement (Dice).}
\label{tab:iaa_per_label}
\end{table*}
Since we use a multi-label annotation schema, we decompose the task into independent binary decisions per label. Table~\ref{tab:iaa_per_label} reports agreement for all labels in the taxonomy.




\paragraph{Pairwise Cohen's $\kappa$.} Computed on flattened binary label vectors, $\kappa$ ranges from 0.747 to 0.819 across all six annotator pairs (0.773--0.850 at the top-level), indicating substantial agreement.

\paragraph{Krippendorff's $\alpha$.} At the top-level, $\alpha$ indicates substantial agreement on categories 0 and 1 ($\alpha \geq 0.67$), while showing moderate agreement on categories 2-5 ($\alpha = 0.54$--$0.66$). Among individual sub-narratives, the highest agreement was observed for 1\_6 (``Sea level rise is exaggerated''; $\alpha = 0.795$), 3\_2 (``Impacts are beneficial''; $\alpha = 0.688$), and 1\_1 (``Ice isn't melting''; $\alpha = 0.672$). Labels in the denial-of-cause family (2\_1, 2\_3, 2\_5) and solutions-won't-work family (4\_2, 4\_4) showed lower agreement ($\alpha = 0.26$--$0.58$), reflecting the greater interpretive difficulty of distinguishing closely related sub-narratives.

\paragraph{Positive Agreement.} For rare labels, $\kappa$ and $\alpha$ can be misleadingly low due to the prevalence paradox: agreement on label absence inflates expected agreement, deflating chance-corrected scores. We therefore also report Positive Agreement (PA), using the Dice coefficient over annotator pairs: 

\[
PA = \frac{2TP}{2TP+ FP + FN}
\]

For the most frequent disinformation labels (2\_1, 5\_1, 2\_3), PA ranges from 0.57 to 0.59, confirming moderate positive-case agreement consistent with Krippendorff's $\alpha$.

\paragraph{Negative Agreement.} Label 2\_2 (``Human greenhouse gases are not causing climate change. It's non-greenhouse gas human climate forcings (aerosols, land use)'') exhibited negative agreement ($\alpha = -0.003$), meaning annotators disagreed more than chance. It was applied 12 times across all annotators, never by more than one annotator on the same item. In each case, the claim discussed topics like deforestation or aerosols but stated factually accurate information; one annotator coded based on topic presence while the others correctly assigned 0\_0. In 23 of the pairwise comparisons, the non-2\_2 annotator assigned 0\_0. The negative agreement thus reflects a systematic confusion between topic and narrative. After adjudication, 2\_2 was retained in only one case.

\paragraph{Annotation Edge Cases.} Three recurring annotation challenges during the annotation process: \begin{enumerate*}[label=\arabic*.]
    \item Mention vs.\ endorsement: claims that reference a disinformation narrative but immediately question it (e.g., ``Some argue that CO$_2$ isn't the main driver of global warming. This is a controversial viewpoint, as the vast majority of scientists agree CO2 plays a significant role in climate change.'') were labeled 0\_0, as the text is \emph{about} the narrative rather than an instance of it; 
    \item Disinformation intent in factually true statements: a claim such as ``Jupiter's weather is powered by a giant internal heat source. \#Climate'' is scientifically accurate, but the \#Climate hashtag hints at an implicit natural-cycles framing (2\_1); annotators disagreed on whether to code surface content or inferred intent, and such cases were resolved conservatively as~0\_0;
    \item Tone-dependent ambiguity: ``Scientists are still working on the final numbers for the projected sea level rise'' can be read as neutral reporting or as implying that climate science is unsettled (5\_1); adjudicators independently converged on~5\_1, interpreting ``final'' as doubting scientific consensus.
\end{enumerate*}

\paragraph{Summary. } 
At the subnarrative level, agreement varies considerably. Labels with clear, observable indicators (e.g., 1\_6: sea level claims; 1\_1: ice/glacier claims) achieve substantial agreement, while labels requiring more interpretive judgment show moderate to fair agreement like within categories~3 (\emph{impacts not bad}) and~4 (\emph{solutions won't work}). This pattern is consistent with the annotation evaluation of the original CARDS taxonomy, where top-level distinctions are more reliable than fine-grained ones \citet{coan2021computer}. The 16.6\% of claims requiring adjudication were resolved by two authors who had not served as annotators, ensuring independence between the annotation and adjudication stages. Each adjudicator first reviewed the claim independently before a joint discussion, and all decisions were documented with reasoning.

\section{Evaluation Metrics}
\label{app:eval-metrics}

Recall@K is defined as:

\[
\text{R@K} =
\frac{\# \text{gold evidentiary abstracts in top-K}}
{\# \text{gold evidentiary abstracts for the claim}}
\]

Because relevance judgments are incomplete, we additionally report Bpref, which does not assume exhaustive annotation~\cite{buckley2004retrieval}, and is defined as:
\[
\text{Bpref} = \frac{1}{R}\sum_{r} \left(1 - \frac{|n \text{ ranked higher than } r|}{R}\right)
\]
where $R$ is the total number of judged relevant documents, $r$ is a judged relevant document retrieved by the system, and $n$ is a member of the first $R$ judged non-relevant documents retrieved before $r$.

\section{\evr Adaptation Details}
\label{app:ev2r}

\subsection{Reference-based Component}
\label{app:atomic_prompt}

\begin{figure*}[h!]
    \centering
    \includegraphics[width=0.85\textwidth]{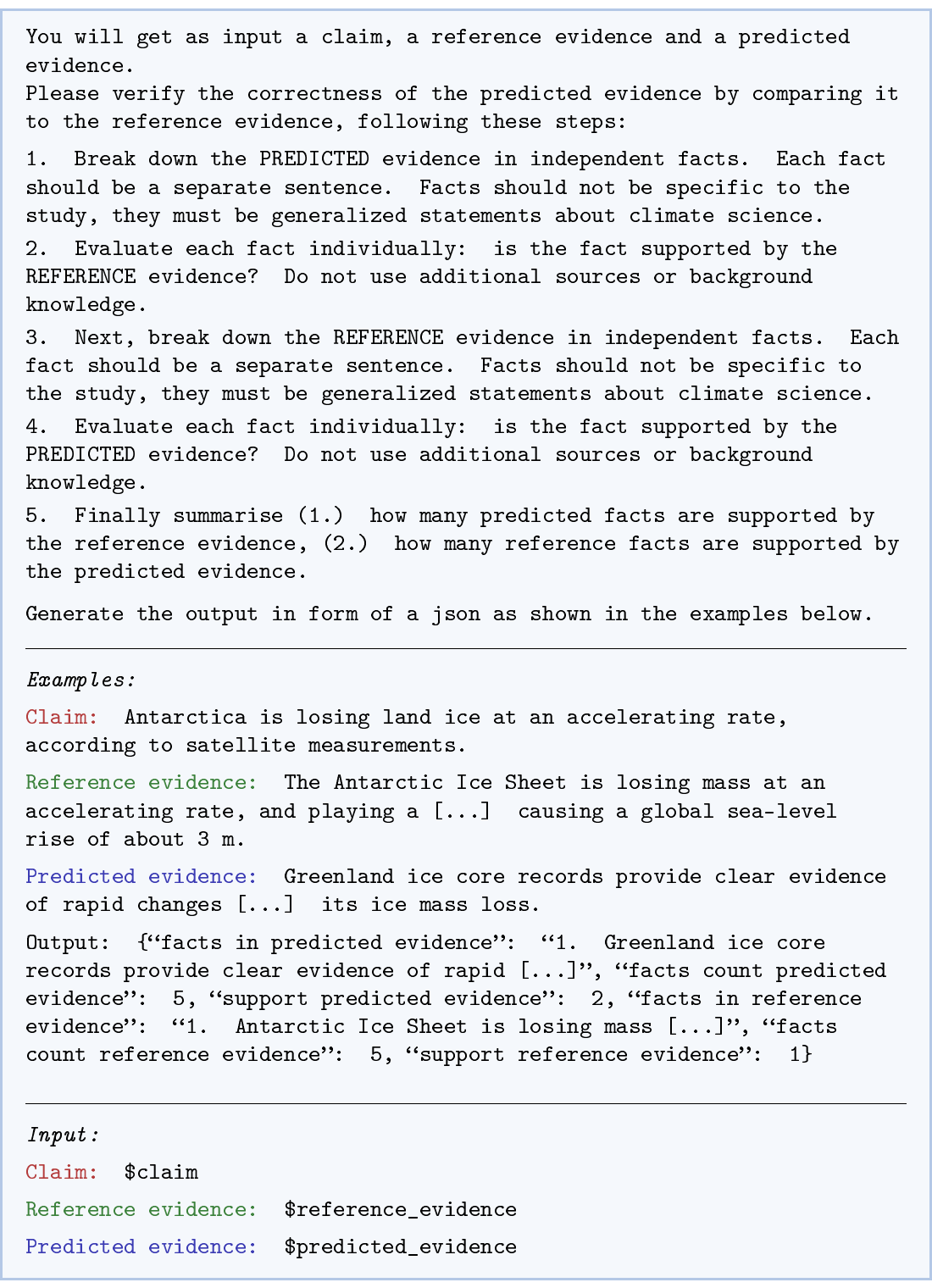}
    \caption{Prompt used for the Ev$^2$R reference-based scorer adapted for the ClimateCheck shared task based on the prompt provided by~\citet{akhtar2024ev2r}. Both the reference evidence and the retrieved evidence are decomposed into atomic facts representing generalised climate science statements before being assessed against each other.}
    \label{fig:ev2r_prompt}
\end{figure*}

To compute the reference-based component of \evr, we use the Gemini 2.5 Pro model~\cite{comanici2025gemini} to decompose gold and retrieved abstracts into atomic factual statements and evaluate their alignment. We adapt the prompt shared by \citet{akhtar2024ev2r} to an instance from the ClimateCheck dataset, shown in Figure~\ref{fig:ev2r_prompt}.\footnote{The full prompt is available at: \url{https://github.com/ryabhmd/climatecheck/blob/master/automatic_eval/reference_based_prompt.txt}} For each retrieved abstract $r$ and gold abstract $g$, the resulting F1 score is used as the reference-alignment score $F1(r,g)$. 
When multiple gold abstracts are available for a claim, we retain the abstract with the the maximum alignment score. If several gold abstracts achieve the same score, the first one is chosen. We cache the results across all submissions to prevent score variations for the same CAP. 

\subsection{Trained DeBERTa Classifier}
\label{app:ev2r_deberta}
To compute the proxy-reference component of \evr, as well as the automatic 
claim verification score for task~1.2, we train a DeBERTa-based classifier 
for 3-way NLI-style claim verification with the labels \emph{Supports}, 
\emph{Refutes}, and \emph{NEI}. 

The model is initialised from an existing classifier,\footnote{\url{https://huggingface.co/MoritzLaurer/DeBERTa-v3-large-mnli-fever-anli-ling-wanli}} a DeBERTa-v3-Large checkpoint pre-trained on a 
diverse set of NLI datasets. We fine-tune it on a combined training set 
drawn from five datasets: FEVER~\cite{Thorne18Fever}, 
VitaminC~\cite{schuster-etal-2021-get}, HoVer~\cite{jiang-etal-2020-hover}, 
AVeriTeC~\cite{schlichtkrull-etal-2024-automated}, and our own ClimateCheck data. For ClimateCheck, we use a stratified 
90/10 train/validation split, ignoring the official test 
split to prevent leakage. The combined training set comprises 488,018 
examples and the validation set 77,843 examples.

Training is performed for 3 epochs on a single NVIDIA H100 GPU with a 
batch size of 32, a learning rate of $2 \times 10^{-6}$, linear learning 
rate scheduling with 200 warmup steps, and a maximum sequence length of 
320 tokens. The best checkpoint is selected by evaluation accuracy on the 
combined validation set, evaluated every 2,000 steps. The selected 
checkpoint at step 26,000 (epoch~1.7) achieves a macro-F1 of 0.886 and 
an accuracy of 0.921 on the combined validation set, with per-class F1 
scores of 0.955 (\textit{supports}), 0.920 (\textit{refutes}), and 0.782 
(\textit{NEI}). We make the model publicly available.\footnote{\url{https://huggingface.co/rausch/deberta-climatecheck-2463191-step26000}}

\section{Disinformation Narrative Classification Prompt}
\label{app:narr_prompt}

The prompt shown in Figure~\ref{fig:narr_prompt} was used for fine-tuning the baseline model for task~2.

\begin{center}
\includegraphics[width=\columnwidth]{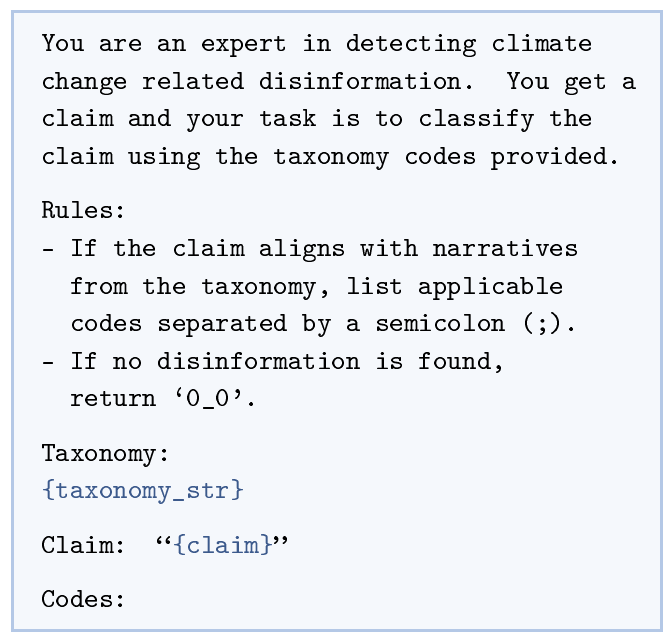}
\captionof{figure}{Prompt used for the baseline implementation of task~2.}
\label{fig:narr_prompt}
\end{center}

\section{Further Error Analyses}
\label{app:error-analysis}
\begin{figure}[th!]
    
    \centering
    \includegraphics[width=0.5\columnwidth]{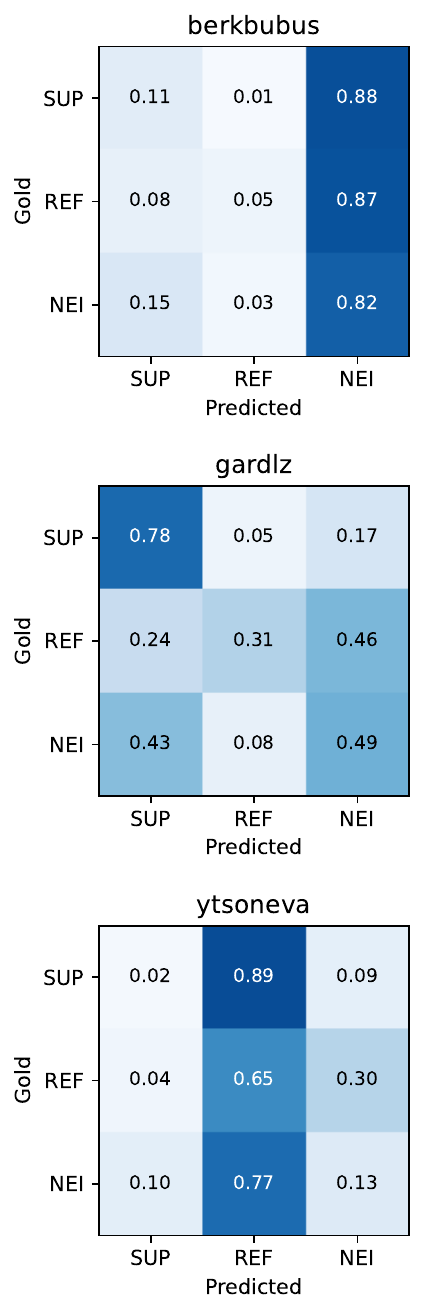}
    \caption{Confusion matrices for the berkbubus, gardlz, and ytsoneva predictions on task~1.2, normalised by claim. SUP = Supports, REF = Refutes, NEI = Not Enough Information.}
    \label{fig:confusion_matrices_2}
\end{figure}

Figure~\ref{fig:confusion_matrices_2} shows the confusion matrices of submitted systems from \emph{berkbubus}, \emph{gardlz}, and \emph{ytsoneva} on task~1.2. Table~\ref{tab:retrieval_by_narrative} displays the R@5 results for the two systems submitting for both tasks per narrative group based on the CARDS taxonomy. No notable differences in retrieval appear to exist, indicating that retrieval difficulty is narrative-agnostic.

\begin{table}[h!]
\small
\center
\begin{tabular}{lrr}
\toprule
 Narrative Group & R@5 & N  \\
\midrule
\multicolumn{3}{@{}c}{\textit{Baseline}} \\
\midrule
0 & 0.992  & 124  \\
1 & 0.933  & 15  \\
2 & 0.938 & 16  \\
3 & 1.000  & 6  \\
4 & 1.000  & 6 \\
5 & 1.000  & 11  \\
\midrule
\multicolumn{3}{@{}c}{\textit{ClimateSense}} \\
\midrule
0  & 0.992  & 124 \\
1  & 1.000  & 15 \\
2  & 1.000  & 16 \\
3  & 1.000  & 6 \\
4  & 1.000  & 6 \\
5  & 0.909  & 11 \\
\bottomrule
\end{tabular}
\caption{R@5 by narrative group for both systems. N indicates the number of claims per group. Results show minimal variance across narrative groups, indicating that retrieval difficulty is narrative-agnostic.}
\label{tab:retrieval_by_narrative}
\end{table}

\end{document}